\documentclass[11pt]{article}
\usepackage{nodalida2021}
\usepackage{times}
\usepackage{url}
\usepackage{latexsym}
\usepackage{adjustbox}
\usepackage[utf8]{inputenc}
\usepackage{todonotes}
\usepackage[estonian, english]{babel}
\usepackage{graphicx}
\usepackage{float}
\usepackage{placeins}
\usepackage{hyperref}

\aclfinalcopy 

\title{Extremely low-resource machine translation for closely related languages}

\author{Maali Tars, Andre Tättar, Mark Fišel\\
University of Tartu\\
{\tt \{maali.tars, andre.tattar, mark.fishel\}@ut.ee}}
\date{}

\begin{document}
\maketitle

\begin{abstract}
  An effective method to improve extremely low-resource neural machine translation is multilingual training, which can be improved by leveraging monolingual data to create synthetic bilingual corpora using the back-translation method. This work focuses on closely related languages from the Uralic language family: from Estonian and Finnish geographical regions. We find that multilingual learning and synthetic corpora increase the translation quality in every language pair for which we have data. We show that transfer learning and fine-tuning are very effective for doing low-resource machine translation and achieve the best results. We collected new parallel data for Võro, North and South Saami and present first results of neural machine translation for these languages.
\end{abstract}

\section{Introduction}

Neural machine translation \citep[NMT,][]{transformer-vaswani} shows great results in terms of output fluency and overall translation quality, however it relies on large parallel corpora for training the models. Low-resource NMT techniques like back-translation \cite{sennrich-bt}, multilingual knowledge transfer \cite{johnson-etal-googles,ngo-etal-2020-improving} and unsupervised NMT \cite{lample-etal-2018-phrase} rely on using parallel corpora for other languages and/or large quantities of monolingual data for the language(s) of interest.

Here we put these techniques to the test in an extremely low-resource setting, working on NMT systems for Võro-Estonian. While Estonian has plentiful parallel, monolingual and annotated corpora \citep[][etc]{tiedemann-2016-opus,universaldeps}, Võro with its 87 000 speakers and no normalized orthography only has slightly over 162 000 monolingual sentences with much less parallel data.

Here we resort to the help of languages closely related to Võro and Estonian: the resource-rich Finnish and two more extremely low-resource North and South Saami. We combine multilingual transfer learning, back-translation and then evaluate several combinations of these techniques on NMT for the five chosen Uralic languages.

Our contributions in this paper are as follows:
\begin{itemize}
    \item experimental results for combinations of techniques for low-resource NMT with application to closely related resource-poor Uralic languages
    \item first developed NMT systems for Võro, North and South Saami languages with a free online demo\footnote{\url{https://soome-ugri.neurotolge.ee/}}
    \item additional data collected for Võro, North and South Saami
\end{itemize}

Next we review related work in Section~\ref{sctRel}, describe our experimental setup in Section~\ref{sctExp}, then proceed with results in Section~\ref{sctRes} and conclude the paper in Section~\ref{sctConc}.

\section{Related work}
\label{sctRel}
This section describes prior work in machine translation (MT) with neural networks for low-resource related languages. Our work on neural machine translation relies on \citep{transformer-vaswani}, who introduce transformer, an encoder-decoder type of solution for MT based on self-attention. 

\subsection{Low-resource NMT}
There has been a lot of research into low-resource MT, for example, phrase-based unsupervised and semi-supervised MT \citep{lample-etal-2018-phrase, artetxe-etal-2018-unsupervised}, but they relied on lexicons or large quantities of monolingual data. Their work is not easily applicable for our experiments because the amount of monolingual data is not sufficient, having less than 100K sentences for most of the languages in our data sets. The authors in \citep{helsinki-low-res-nmt} used a template based approach to generate more parallel data for related languages, which made NMT models viable for training.

Another way of doing multilingual NMT is via zero-shot translations for very low resource language pairs. In our case, we have data for ten translation directions and zero parallel data for the rest of the ten directions. In \citep{fb-low-nmt} the authors showed that zero-shot translations achieve better results than the pivoting approach - pivoting means that when we have a language pair with sufficient data, then Võro to Finnish translation, which has zero data, would use the Estonian language to pivot - Võro to Estonian to Finnish translation. We want to avoid pivoting because Võro to North Saami would require two pivots or three translations in total, resulting in serious error propagation. Additionally, the authors use shared source embeddings and source RNN encoders; we used transformers with shared vocabulary, encoders and decoders. In \citep{rikters-etal-2018-training} the authors showed that multilingual training with transformers is optimal for multilingual Estonian-English-Russian system, but reported that high-resource pairs see a performance degradation and lower-resourced pairs see a performance increase. 

\subsection{Back-translation for low-resource MT}
Every sentence is essential for neural machine translation in a low-resource machine translation environment. One popular way to leverage monolingual data is by creating a synthetic corpus via a method called back-translation (BT). Traditional BT \cite{sennrich-bt} is easy to use and requires training a target-to-source MT system to generate translations of the monolingual data, which are used as training data for the source-to-target MT model. This means that traditional BT requires two NMT models, where one generates synthetic data for the other. The idea behind BT is that the monolingual human data on the target side improves the quality of the decoder to generate better output for the language and the synthetic source helps as a data augmentation tactic.

Closely related to back-translation is a method called forward-translation (FT), where the model creates synthetic parallel data for itself - the source sentence is translated into the target language, and together, a bitext sample is created. In other words, forward-translation is called self-training. The authors in \citep{popovic-etal-2020-neural} used both BT and FT for closely related languages. They used a multilingual encoder (English and German) and a multilingual decoder (Serbian and Croatian) and achieved better results compared to single directional baselines in their experiments.

Our work is about a single multilingual system that enables the model to generate synthetic data for itself - both back-translation and forward-translation is used. The generated synthetic data is added to available parallel corpora as training data.

\subsection{Transfer learning and fine-tuning}

The authors in \citep{kocmi-bojar-2018-trivial} did trivial transfer learning for low resource NMT - in detail, they used a high resource language pair like English-Finnish to train a parent model. They continued training on a lower resource child model English-Estonian and showed that this improved translation quality significantly, 19.74 BLEU score compared to 17.03 when using only English to Estonian data. Additionally, they showed that ``unrelated'' languages might work even better, where the best English-Estonian results were achieved by using an English-Czech as a parent, which achieved a 20.41 BLEU score on the same test set. Their work shows that transfer learning is a very viable option for low-resource NMT. The only drawback is that their work still relies on some amount of data and a common source or target language to either share the encoder or decoder weights. In our case, there are language pairs, which have 0 available sentences like Võro to North Saami. Additionally, their work would require 20 such models to be trained.

The authors in \citep{currey-heafield-2019-zero, zhang-etal-2020-improving} show that using multilingual back-translation for fine-tuning a multilingual model is beneficial for translation quality. Additionally, \citep{zhang-etal-2020-improving} shows that their random online back-translation lowers the chance of the model doing off-target translations, which in our case is also a problem since the model never sees some language pairs. We build upon this work by doing two iterations of fine-tuning on a synthetic back-translation corpora, where we uniformly at random assign the target language into which to translate.

The difference between transfer learning and fine-tuning is small. We refer to transfer learning when the MT model is trained on some languages that the model has never seen before, e.g., when using ET-FI model weights to initialize the ET-VRO model.  We refer to fine-tuning when we continue training a multilingual model on data, which the model has seen before, e.g., when using the multilingual model to fine-tune on ET-VRO data only.

\subsection{Source factors for multilingual zero-shot NMT}
We rely on the work of \citep{sennrich-haddow-2016-linguistic} for zero-shot translations in our multilingual models; in that article, the authors use morphological features like POS tags to enrich source-side representations. We use source-side factors to give the transformer model information about the intended target language, so the model knows which language the output should be in. The authors in \citep{tars-fishel-2018} used source-factors to give domain and target language information for the model.
Using source factors is similar to using a single token on the input sentence to distinguish between closely related languages and dialects \citep{lakew-etal-2018-neural, costa-jussa-etal-2018-neural}, where authors show an improvement over a single baseline model when training a model for similar languages.

\begin{table*}
\centering
\begin{tabular}{ l|r|r|r } 
 \hline
 Language pair & Before cleaning & After cleaning & Eliminated \\
 \hline\hline
 et-fi \citep{tiedemann-2016-opus} 
 & 3 566 826 & 2 646 922 & 919 904 \\
 \hline
 et-vro \footnotemark[2] 
 & 30 816 & 30 502 & 314 \\
 \hline
 fi-sme \footnotemark[3] 
 & 109 852 & 35 426 & 74 426 \\ 
 \hline
 fi-sma  \footnotemark[3] 
 & 3098 & 2895 & 203 \\
 \hline
 sme-sma  \footnotemark[3] 
 & 23 746 & 21 557 & 2189 \\ 
 \hline\hline
 Overall & 3 734 338 & 2 737 302 &  997 036 \\
 \hline
\end{tabular}
\caption{Parallel data sets (in sentence pairs). et - Estonian, fi - Finnish, vro - Võro, sme - North Saami, sma - South Saami.}
\label{table:1}
\end{table*}

\section{Experimental setup}
\label{sctExp}

\subsection{Data sets}

\subsubsection{Preprocessing}

The data for the experiments originated from many different sources. Subsequently, the main issue with the parallel data collected was the differences in file formats, which took a long time to solve in order to create a unified data set. The biggest problem with parallel data was that there were a lot of repeated sentence pairs in the data, which required a uniqueness check and reduced the number of sentence pairs for the Finnish-North Saami (FI-SME) language pair by about 75 percent, as seen in Table \ref{table:1}. 

Preprocessing monolingual data was also a long process as there were no conclusive ready-made sets available for languages like Võro, North Saami and South Saami. As described in Table \ref{table:2}, in the first set, the data consisted mostly of news corpuses, fiction and Wikipedia texts. The data files were in different formats, as was the case with parallel data. Estonian and Võro required extracting sentences from texts and removing empty lines. The Võro, North Saami and South Saami data in the second set was gathered manually from news articles and various PDF style documents (fiction, scientific texts, official documents) available. The paragraphs of text then needed to be divided into sentences and joined into one TXT type file for compatibility. Additional preprocessing included fixing some minor alignment issues. 


\subsubsection{Validation and test data}

Validation and test sets consisted of sentences from all the five language pairs mentioned in Table \ref{table:1}. The number of sentences for each language pair was chosen proportionally to the amount of training data the pair had. In total, there were 1862 test sentences and 939 validation sentences. There is no official test set available for these language pairs collectively, and as parallel data was scarce, the validation and test sets were sentences that were randomly held-out of the training data.

\subsubsection{Parallel data}

Table \ref{table:1} also highlights the fact that Estonian-Finnish (ET-FI) acted as the high-resource language pair in the experiments, with 2.6 million sentence pairs available. Other language pairs formed a small fraction of the whole parallel data set, with about 1 percent. The lowest amount of data was discovered for Finnish-South Saami (FI-SMA) language pair, with under 3000 sentence pairs.

\subsubsection{Monolingual data}

As expected, Estonian and Finnish had the most monolingual data available. Although finding data sets for the low-resource languages proved to be more difficult, there was more of it available than parallel data for their respective language pairs used in this work. The two sets of monolingual data described in Table \ref{table:2} were collected separately. Experiments with the first set were already performed prior to gathering the second set, which is why the amounts of two sets are off-balance.
For the sake of the models learning more about low-resource languages, we used the down-sampling technique, reducing the amount of Estonian and Finnish monolingual data to level them with the amount of low-resource language monolingual data in use. That made Võro (VRO) the most prominent language in the data set, as shown in Table \ref{table:2}. The amount of data for North and South Saami was still quite low, but it was an improvement over the parallel data set numbers.

\footnotetext[2]{\url{https://doi.org/10.15155/1-00-0000-0000-0000-001A0L}}
\footnotetext[3]{\url{https://giellalt.uit.no/tm/TranslationMemory.html}}

\begin{table*}
\centering
\begin{tabular}{ l|r|r|r } 
 \hline
 Language & First set & Second set & All  \\
 \hline\hline
 et \footnotemark[4] 
 & 100 000 & 25 000 & 125 000 \\
 \hline
 fi \citep{wortschatzpaper}
 & 100 000 & 25 000 & 125 000 \\
 \hline
 vro \footnotemark[5],\footnotemark[6]
 & 162 807 & 5290 & 168 097 \\ 
 \hline
 sme \citep{wortschatzpaper,TIEDEMANN12.463},
 \footnotemark[6]
 & 33 964 & 6057 & 40 021 \\
 \hline
 sma\footnotemark[6],\footnotemark[7] \citep{TIEDEMANN12.463},
 & 55 088 & 5377 & 60 465 \\ 
 \hline
\end{tabular}
\caption{Monolingual data sets after preliminary cleaning (in sentences). et - Estonian, fi - Finnish, vro - Võro, sme - North Saami, sma - South Saami.}
\label{table:2}
\end{table*}
\footnotetext[4]{\url{https://www.cl.ut.ee/korpused/segakorpus/epl/}}
\footnotetext[5]{\url{https://doi.org/10.15155/1-00-0000-0000-0000-00186L}}
\footnotetext[6]{\url{https://github.com/maalitars/FinnoUgricData}}
\footnotetext[7]{\url{http://hdl.handle.net/11509/102}}

\subsection{Models and parameters}

\subsubsection{General settings}

In our experiments we use the Sockeye framework described by \citep{sockeye}, which has implemented source-side factors where we give the target language token as an input feature for the transformer model. During training, the vocabulary that was created included all of the languages. Specifications of the training process included setting the batch size to 6000 words and checkpoint interval to 2000. All the models in the experiments trained until 32 consecutive unimproved checkpoints were reached. The unimproved metric was perplexity. All of the experiments use the standard transformer parameters (6 encoder and 6 decoder layers with 8 attention heads and size 512).
Prior to training, all of the data used to develop the models was tokenized by a SentencePiece \citep{kudo-richardson-2018-sentencepiece} tokenization model, which follows the byte-pair encoding algorithm. The tokenization model was previously trained on all of the training data. 

\subsubsection{Multilingual baseline}

One of the fundamental experiments of this work was developing the multilingual baseline model, which had five source languages and five target languages. This means that this model could produce translations in 20 different directions. For this, each pair of parallel data seen in Table \ref{table:1} was copied and the source-target direction was switched. The turned-around parallel data set was then added to the original data set and the multilingual baseline model was trained on all of the combined parallel data.
The data set was tokenized by a tokenization model, which was trained on all the training data from the parallel data set, meaning text patterns were generalized over five languages.

\subsubsection{Back-translation experiments}

Synthetic parallel data via back-translation was produced in two iterations and additional models were also trained in two iterations. The monolingual data was translated into every other language in equal measures. For example, 1/4 of the 100 000 sentences in Estonian were translated into Finnish, 1/4 into Võro, 1/4 into North Saami and 1/4 into South Saami. The paired-up synthetic translations and monolingual data made up the additional parallel data corpus.

\begin{table*}[t]
\centering
\begin{adjustbox}{max width=1.0\textwidth}
\begin{tabular}{ l|c|c|c|c|c|c|c|c|c|c|c } 
 Model & et-fi & fi-et & et-vro & vro-et & fi-sme & sme-fi & fi-sma & sma-fi & sme-sma & sma-sme & $BLEU_{low}$  \\
 \hline\hline
 Baselines & 32.0 & 29.4 & 14.6 & 17.5 & 28.0 & 28.7 & 4.6 & 6.3 & 8.3 & 9.1 & 14.6\\
 \hline\hline
 Multilingual (ML) & 30.9 & 29.5 & 23.8 & 29.6 & 31.3 & 34.7 & 9.4 & 9.4 & 19.8 & 19.8 & 22.2\\
 \hline\hline
 + BT1 & \textbf{32.4} & 29.9 & 25.2 & 29.4 & \textbf{32.3} & 36.1 & 10.8 & 9.9 & 20.3 & 20.0 & 23.0 \\ 
 + BT1(*) & 30.1 & 29.1 & 24.5 & 30.3 & \textbf{32.3} & 36.2 & 11.1 & 10.5 & 21.4 & 20.0 & 23.3 \\
 + BT1 + FT1 & 31.3 & 30.1 & 25.2 & 31.5 & 31.3 & 35.7 & 8.9 & 10.0 & 18.7 & 20.4 & 22.7 \\
 + BT1 + FT1(*) & 30.9 & 28.8 & 25.8 & 30.4 & 31.5 & 35.7 & 8.9 & 10.1 & 19.4 & 20.1 & 22.7 \\
 \hline\hline
 + BT2 & 31.5 & \textbf{30.2} & 26.0 & 31.0 & \textbf{32.3} & 36.6 & 11.3 & \textbf{10.9} & 20.3 & \textbf{21.0} & 23.7 \\
 + BT1 + BT2(*) & 31.3 & 29.6 & \textbf{26.2} & 31.3 & 31.4 & 36.4 & \textbf{12.4} & 10.6 & \textbf{21.6} & 20.7 & \textbf{23.8} \\
 + BT1 + BT2(**) & 30.4 & 29.7 & 25.1 & 31.6 &	31.7 & \textbf{37.5} &	11.4 & 10.3 & 21.3 & 20.9 & 23.7 \\
 + BT1\&2 + FT1\&2(*) & 30.2 & 29.4 & 25.1 & \textbf{31.7} & 31.5 & 36.8 & 9.5 & 9.7 & 20.4 & 20.6 & 23.2 \\
 \hline\hline
 BT1 & 21.1 & 21.6 & 20.5 & 24.9 & 24.0 & 27.4 & 8.5 & 7.3 & 15.9 & 14.5 & 17.9 \\
 BT1(*) & 8.4 & 8.6 & 18.7 & 19.9 & 11.8 & 13.4 & 6.9 & 5.3 & 12.9 & 9.2 & 12.3 \\
\end{tabular}
\end{adjustbox}
\caption{BLEU scores. (*) - trained without pre-trained weights, (**) - trained on \textit{+ BT1(*)} weights. \textit{BT} - back-translation data set, \textit{FT} - forward-translation data set, $BLEU_{low}$ - average BLEU score on low-resource language pairs (excluding ET-FI and FI-ET), \textbf{bold} - best BLEU score for a language pair.}
\label{table:3}
\end{table*}

Combining the new synthetic parallel data corpus and the original, human-translated corpus, gives the models more parallel data to learn on during training. The methodology of both back-translation data experiment iterations was the same, but there were some important aspects that were different:

\textbf{First iteration}
\begin{itemize}
\item Monolingual data used: first monolingual data set 
\item The first batch of synthetic data was produced with the multilingual baseline model. The synthetic data was then added to the original parallel data and the training process was repeated, which produced a new model. 
\end{itemize}

\textbf{Second iteration}
\begin{itemize}
\item Monolingual data used in this iteration consisted of 1) shuffled first monolingual data set, 2) second monolingual data set.
\item Monolingual data was translated by the newest model that had been trained on parallel data and synthetic data from the first iteration of back-translation (\textit{+BT1} in Table \ref{table:3}). Subsequently, a new model was trained using original parallel data plus the two batches of synthetic data produced.
\end{itemize}
Additional experiments included having different combinations of back-translation/forward-translation data and differences in initialized weights, with the best of them presented in Table \ref{table:3}.

\subsubsection{Transfer learning experiments}

We performed an experiment fine-tuning the multilingual baseline model on ET-VRO parallel data and a transfer learning experiment, initializing ET-VRO model with ET-FI baseline model weights. Then we compared the results of these two experiments to each other and to the ET-VRO baseline model. The ET-VRO data was the same parallel data that was used for training the multilingual baseline model (\textit{ML}).

\section{Results}
\label{sctRes}

\subsection{Quantitative analysis}
Quantitative results were determined by comparing BLEU scores \citep{papineni-etal-2002-bleu}, using the SacreBLEU implementation \citep{post-2018-call} of calculating the score on detokenized sentences\footnotemark[8]. Multiple experiments were assessed and the best experiments are explained in Table \ref{table:3}. Additional analysis was done with the CHRF metric, which compares sentences on a character-level \citep{popovic-2015-chrf}. We used the SacreBLEU implementation \citep{post-2018-call} of the CHRF metric\footnotemark[9]
 and the results can be seen in the Appendix in Table \ref{table:9}.

\subsubsection{BLEU}

\textbf{Multilingual baseline.} All of the low-resource language pairs experienced a positive gain over baseline model results in comparison to the multilingual baseline model (\textit{ML}) experiment. An average gain of 7.6 BLEU was achieved on the low-resource language pairs with VRO-ET and SME-SMA exceeding this average gain by an additional 4 BLEU points. Noticeably, FI-SMA made the smallest improvement, perhaps the main reason for this lies in FI-SMA having significantly less parallel data than other low-resource language pairs.

\footnotetext[8]{SacreBLEU signature: BLEU+case.mixed+lang.LANG-LANG+numrefs.1+smooth.exp+test.SET+tok.13a+\linebreak version.1.4.14 where LANG in \{et,fi,vro,sme,sma\}}

\footnotetext[9]{SacreBLEU signature: 
chrF2+lang.LANG-LANG+numchars.6+numrefs.1+space.false+test.SET+
\linebreak version.1.5.1 where LANG in \{et,fi,vro,sme,sma\}}

\begin{table*}
\centering
\begin{tabular}{ l l p{10cm} } 
\hline
a) & Source & Uue nime \textbf{väljamõtlemisel} oli tähtis, et oleks selge side kohaliku kogukonnaga ja et nimi \textbf{aitaks jutustada} ettevõtte lugu. \\
\hline
& Baseline & Vahtsõ opimatõrjaali saamisõs oll tähtsä, et tähtsä olõs ka selge sõnumiga tõsitsit luulõtuisi. \\
& ML & Vahtsõ nime \textcolor{red}{vällämõtlemisel} oll tähtsä, et olõsi selge side paigapäälitse kogokunnaga ja et nimi \textcolor{red}{avitas kõnõlda} ettevõtte lugu. \\
& +BT1+BT2(*) & Vahtsõ nime \textcolor{red}{vällämõtõldõn} oll’ tähtsä, et olõs selge side paigapäälidse kogokunnaga ja et nimi \textcolor{red}{avitas kõnõlda} ettevõttõ lugu. \\
\hline
& Reference & Vahtsõ nime \textbf{vällämärkmise man} oll’ tähtsä, et olõs selge köüdüs paikligu kogokunnaga ja et nimi \textbf{avitanuq jutustaq} ettevõtmisõ luku. \\
& \textit{English} & On \textbf{coming up with} a new name, it was important that there was a clear reference to the local community and that the name would \textbf{help tell the story} of the business.\\ 
\hline
\\
\\
\hline
b) & Source & Parhilla ommaq jutuq hindamiskogo käen, kokko\textbf{võtõq} ja preemiäsaajaq trüki\textbf{täseq} ärq järgmädsen \textbf{Uman Lehen}.\\
 \hline
 & Baseline & Praegu on instruktsioone, ka kokku\textcolor{red}{võtted}, selliste sündmuste ja aegajalt „sisse lülitada” kaugemate Leivalentsemad.\\
 & ML & Praegu on jutud hindamiskogu käes, kokku\textcolor{red}{võtted} ja preemiasaajad trüki\textcolor{red}{vad} järgmise \textcolor{red}{Uman Leheni}.\\
 & +BT1\&2+FT1\&2(*) & Praegu on jutud hindamiskogu käes , kokku\textcolor{red}{võtted} ja preemiasaajad trüki\textcolor{red}{vad} ära järgmises \textcolor{red}{Uman Lehes} .\\
 \hline
 & Reference & Praegu on jutud hindamiskomisjoni käes, kokku\textbf{võte} ja preemiasaajad trüki\textbf{takse} ära järgmises \textbf{Uma Lehes}.\\
 & \textit{English} & At the moment the stories are with the judging committee, \textbf{the summary} and the winners will be \textbf{printed} in the next \textbf{Uma Leht}. \\
 \hline
 \\
 \\
 \hline
c) & Source & Nuoria on tullut tilalle \textbf{aika lailla}, ehkä ottaa eräs nuori kyläelämän vetämisen haltuunsa. \\
\hline
& Baseline & Nuorat leat boahtán sadjái áigi, soađi, kántorin jos čađahat gilvun. \\
& ML & Nuorat leat boahtán sadjái \textcolor{red}{áiggi ládje}, soaitá váldit ovtta nuorra gilieallima jođiheami. \\
& +BT1 & Nuorat leat boahtán sadjái \textcolor{red}{áige ládje}, soaitá váldit ovtta nuorra gilieallima jođiheami háldui. \\
\hline
& Reference & Nuorat lea boahtán lasi \textbf{oalleláhkái}, gánske muhtun nuorra váldá gilieallima geassima iežas háldui. \\
& \textit{English} & There are \textbf{quite a bit} more younger people now, maybe one of them will take over the role of leading the village life. \\
\hline
\end{tabular}
\caption{Example translations from a) ET-VRO, b) VRO-ET and c) FI-SME.}
\label{table:4}
\end{table*}

\textbf{Back-translation experiments.} Experiments with data from back-translation iterations further improved the BLEU score for low-resource language pairs compared to the multilingual baseline model. None of the models showed uniform improvements across all of the low-resource language pairs, however we can highlight one model with the highest average gain over baseline results, improving by +9.2 BLEU points. This model was trained on parallel data plus two batches of back-translation data but without any initialized weights (\textit{+ BT1 + BT2(*)} in Table \ref{table:3}).

While the pre-trained weights did not seem to help produce the best models with parallel and back-translation data, the experiments with only back-translation data show that initializing a model with useful pre-trained weights can still be very helpful in the case of related tasks. This is illustrated by models \textit{BT1} and \textit{BT1(*)} with 7.1 BLEU points between them.

Experiments with added forward-translations did not appear to improve the results except for the VRO-ET language pair.

\textbf{Transfer learning experiments.}
Transfer learning and fine-tuning a model for a particular language pair results in further improvements over the best back-translation model results. In this part, we performed two experiments. In the transfer learning experiment, we trained an ET-FI baseline model until convergence; then the training data was changed to the ET-VRO data set, which was used for training until convergence. In the second experiment, we fine-tuned the multilingual baseline model with the ET-VRO language direction data only. Comparing BLEU results in Table \ref{table:8}, it is clear that doing transfer learning for low-resource NMT is very beneficial - a 12 BLEU point increase is achieved by doing trivial transfer learning, and even better gains are seen in the multilingual fine-tuning experiment with a 13 point BLEU score increase.

Thus, the best results were achieved in the transfer learning and the fine-tuning experiment. Transfer learning alone, however, has a down-side. Compared to multilingual models, which can translate in 20 different directions, in case of transfer learning, to achieve the same functionality, 20 separate models would have to be trained, which takes up a lot more resources.

\subsubsection{CHRF}

For the low-resource language pairs, the CHRF score metric mostly agreed with the BLEU score metric on which model gives the best results for each language pair, except for SMA-FI and SMA-SME. This can be seen in the Appendix in Table \ref{table:9}.
With the CHRF score, however, it is much clearer that the model \textit{+ BT1 + BT2(*)} is the best one out of all the experiments done with back-translations, because both $BLEU_{low}$ and $CHRF_{low}$ had the best scores on test data with this model and six out of the eight low-resource language pairs achieved the highest CHRF scores.
In Table \ref{table:8}, for transfer learning and fine-tuning experiments, the BLEU and CHRF scores moved in the same direction, transfer learning and fine-tuning improving results substantially.\\
%

Overall, we can see the same patterns, both in the BLEU and the CHRF score analysis: the multilingual model concept helps get better translation quality for low-resource languages compared to baseline results; adding more and more back-translated data to the training data increases the scores; adding forward-translations, however, mostly lowers the scores. Another noticeable thing shown by both of the scores, is that back-translation on its own, when looking at the models \textit{BT1} and \textit{BT1(*)}, does not achieve good (and comparable) results. One possible reason for this is the data domain mismatch between test data (parallel data hold-out) and monolingual data. A balanced test set for these languages could provide a better overview of the results and give more accurate info on the quality of the models.

\begin{table}
\centering
\begin{adjustbox}{max width=1.0\textwidth}
\begin{tabular}{l|c|c}
Model & BLEU & CHRF \\
\hline\hline
ET-VRO baseline & 14.6 & 0.393\\
\hline
ET-VRO on ET-FI weights & 26.5 & 0.540\\
\hline
ML fine-tuned on ET-VRO & 27.6 & 0.563\\
\end{tabular}
\end{adjustbox}
\caption{BLEU and CHRF scores for transfer learning and fine-tuning experiments.}
\label{table:8}
\end{table}

\subsection{Qualitative analysis}

Table \ref{table:4} compares some sentence translations for ET-VRO, VRO-ET and FI-SME language pairs. It is clear that baseline models produced subpar translations, rendering them non-sensical. The multilingual baseline model improved the translations significantly, although still making some detrimental mistakes, like choosing the wrong word, so the meaning is lost, or deciding not to translate some parts of the sentence. Adding back-translation data to the models fixed some of these mistakes and made some important changes in understanding the meaning of a sentence, but the best back-translation models still left in some grammatical errors, such as wrong verb forms, grammatical cases and tenses.

In the first example, translating in the ET-VRO direction, the best model chooses a direct translation of ``väljamõtlemisel'' to ``vällämõtõldõn''. In addition, all of the models omit the ``q'' endings of the words ``avitanuq'' and ``jutustaq'' or ``kõnõldaq''. This symbol usually signifies plurality and, in a lot of cases upon translating in the ET-VRO direction, the models chose not to add the ``q'', although it would have been correct. The problem could lie in the data, where the ``q'' endings are also not always added, which in turn could confuse the models.

The second example illustrates a VRO-ET direction translation, which presents some bigger flaws. For example, handling names is difficult even for the best model, with ``Uman Lehen'' being translated incorrectly to ``Uman Lehes''. The grammatical case of the word ``Lehen'' was correctly changed to have an ``s'' ending, but the case of the word ``Uman'' was not changed.
In addition, ``trükitäseq'' is translated to the wrong verb form ``-vad'', but it should be replaced with the impersonal form ``-takse''. Continuing with the problems caused by the ``q'' ending, here the word ``kokkovõtõq'' is translated to plurality ``kokkuvõtted'', but in this particular case, it should be translated to the singular form ``kokkuvõte''.

The third example shows the FI-SME language pair translations. Here the meaning of the sentence is understandable, but the word-pair ``áige ládje'' is a direct translation from the Finnish phrase ``aika lailla'', which means ``quite a bit'' in English, but ``áige ládje'' does not hold the same meaning.

Additional examples can be seen in the Appendix in Table \ref{table:5}. In these examples, there is another flaw presented, which might be unique to multilingual models, where some words in a sentence are translated into the wrong language, although they might have the correct meaning. This is illustrated very well in the third ET-VRO translation example in Table \ref{table:5}. All models, except the baseline, choose to translate the word ``ametlikult'' into the Finnish word ``virallisesti'', instead of trying to find a word for it in Võro language.

\subsection{Discussion}
The results show that synthetic data helps to learn a better model; however, the model which has continued training on only back-translation data sees performance degradation. This is most likely caused by the test sets' domain mismatch problem and is alleviated by merging the parallel and synthetic data into one big corpus. This shows that a new separate test set should be created for this problem, but it is very hard to do as there are very few speakers. We have started to gather a multilingual five-way test corpus.

We think that this work can be further improved by doing better multilingual fine-tuning, shown by promising multilingual fine-tuning experiments, where the best result was 27.6 BLEU points compared to the best multilingual model, which achieved 26.2 BLEU points. The research question remains, how well would fine-tuning work for a completely synthetic parallel corpus like VRO-SMA. We did not explore this yet due to not having enough resources. Also, it is unknown what effect this single language pair fine-tuning would have on other languages.

\section{Conclusions and future work}
\label{sctConc}

Multilingual neural machine translation with shared encoders and decoders work very well for very low resource language translation. Using back-translation for low resource MT is vital for best results, further improved by transfer learning and fine-tuning. 

In the future, we hope to work with more Uralic languages and add an unrelated high-resource language, for example German. Secondly, we want to do better multilingual fine-tuning since the best ET-VRO score of 27.6 was reached by multilingual fine-tuning, compared to 26.2 for multilingual training. Finally, we hope to find more parallel and monolingual data.

\bibliographystyle{acl_natbib}
\bibliography{nodalida2021}

\begin{thebibliography}{26}
\expandafter\ifx\csname natexlab\endcsname\relax\def\natexlab#1{#1}\fi

\bibitem[{Artetxe et~al.(2018)Artetxe, Labaka, and
  Agirre}]{artetxe-etal-2018-unsupervised}
Mikel Artetxe, Gorka Labaka, and Eneko Agirre. 2018.
\newblock \href {https://doi.org/10.18653/v1/D18-1399} {Unsupervised
  statistical machine translation}.
\newblock In \emph{Proceedings of the 2018 Conference on Empirical Methods in
  Natural Language Processing}, pages 3632--3642, Brussels, Belgium.
  Association for Computational Linguistics.

\bibitem[{Costa-juss{\`{a}} et~al.(2018)Costa-juss{\`{a}}, Zampieri, and
  Pal}]{costa-jussa-etal-2018-neural}
Marta~R Costa-juss{\`{a}}, Marcos Zampieri, and Santanu Pal. 2018.
\newblock \href {https://www.aclweb.org/anthology/W18-3931} {{A Neural Approach
  to Language Variety Translation}}.
\newblock In \emph{Proceedings of the Fifth Workshop on {NLP} for Similar
  Languages, Varieties and Dialects ({V}ar{D}ial 2018)}, pages 275--282, Santa
  Fe, New Mexico, USA. Association for Computational Linguistics.

\bibitem[{Currey and Heafield(2019)}]{currey-heafield-2019-zero}
Anna Currey and Kenneth Heafield. 2019.
\newblock \href {https://doi.org/10.18653/v1/D19-5610} {Zero-resource neural
  machine translation with monolingual pivot data}.
\newblock In \emph{Proceedings of the 3rd Workshop on Neural Generation and
  Translation}, pages 99--107, Hong Kong. Association for Computational
  Linguistics.

\bibitem[{Goldhahn et~al.(2012)Goldhahn, Eckart, and
  Quasthoff}]{wortschatzpaper}
D.~Goldhahn, T.~Eckart, and U.~Quasthoff. 2012.
\newblock Building large monolingual dictionaries at the leipzig corpora
  collection: From 100 to 200 languages.
\newblock In \emph{Proceedings of the 8th International Language Resources and
  Evaluation (LREC'12)}.

\bibitem[{Gu et~al.(2018)Gu, Hassan, Devlin, and Li}]{fb-low-nmt}
Jiatao Gu, Hany Hassan, Jacob Devlin, and Victor~O.K. Li. 2018.
\newblock \href {https://doi.org/10.18653/v1/N18-1032} {Universal neural
  machine translation for extremely low resource languages}.
\newblock In \emph{Proceedings of the 2018 Conference of the North {A}merican
  Chapter of the Association for Computational Linguistics: Human Language
  Technologies, Volume 1 (Long Papers)}, pages 344--354, New Orleans,
  Louisiana. Association for Computational Linguistics.

\bibitem[{H\"{a}m\"{a}l\"{a}inen and Alnajjar(2019)}]{helsinki-low-res-nmt}
Mika H\"{a}m\"{a}l\"{a}inen and Khalid Alnajjar. 2019.
\newblock \href {https://doi.org/10.1145/3377713.3377801} {A template based
  approach for training nmt for low-resource uralic languages - a pilot with
  finnish}.
\newblock In \emph{Proceedings of the 2019 2nd International Conference on
  Algorithms, Computing and Artificial Intelligence}, ACAI 2019, page
  520–525, New York, NY, USA. Association for Computing Machinery.

\bibitem[{Hieber et~al.(2017)Hieber, Domhan, Denkowski, Vilar, Sokolov,
  Clifton, and Post}]{sockeye}
Felix Hieber, Tobias Domhan, Michael Denkowski, David Vilar, Artem Sokolov, Ann
  Clifton, and Matt Post. 2017.
\newblock Sockeye: A toolkit for neural machine translation.

\bibitem[{Johnson et~al.(2017)Johnson, Schuster, Le, Krikun, Wu, Chen, Thorat,
  Vi{\'e}gas, Wattenberg, Corrado, Hughes, and Dean}]{johnson-etal-googles}
Melvin Johnson, Mike Schuster, Quoc~V. Le, Maxim Krikun, Yonghui Wu, Zhifeng
  Chen, Nikhil Thorat, Fernanda Vi{\'e}gas, Martin Wattenberg, Greg Corrado,
  Macduff Hughes, and Jeffrey Dean. 2017.
\newblock {G}oogle{'}s multilingual neural machine translation system: Enabling
  zero-shot translation.
\newblock \emph{Transactions of the Association for Computational Linguistics},
  5:339--351.

\bibitem[{Kocmi and Bojar(2018)}]{kocmi-bojar-2018-trivial}
Tom Kocmi and Ond{\v{r}}ej Bojar. 2018.
\newblock \href {https://doi.org/10.18653/v1/W18-6325} {Trivial transfer
  learning for low-resource neural machine translation}.
\newblock In \emph{Proceedings of the Third Conference on Machine Translation:
  Research Papers}, pages 244--252, Brussels, Belgium. Association for
  Computational Linguistics.

\bibitem[{Kudo and Richardson(2018)}]{kudo-richardson-2018-sentencepiece}
Taku Kudo and John Richardson. 2018.
\newblock \href {https://doi.org/10.18653/v1/D18-2012} {{S}entence{P}iece: A
  simple and language independent subword tokenizer and detokenizer for neural
  text processing}.
\newblock In \emph{Proceedings of the 2018 Conference on Empirical Methods in
  Natural Language Processing: System Demonstrations}, pages 66--71, Brussels,
  Belgium. Association for Computational Linguistics.

\bibitem[{Lakew et~al.(2018)Lakew, Erofeeva, and
  Federico}]{lakew-etal-2018-neural}
Surafel~Melaku Lakew, Aliia Erofeeva, and Marcello Federico. 2018.
\newblock \href {https://doi.org/10.18653/v1/W18-6316} {{Neural Machine
  Translation into Language Varieties}}.
\newblock In \emph{Proceedings of the Third Conference on Machine Translation:
  Research Papers}, pages 156--164, Brussels, Belgium. Association for
  Computational Linguistics.

\bibitem[{Lample et~al.(2018)Lample, Ott, Conneau, Denoyer, and
  Ranzato}]{lample-etal-2018-phrase}
Guillaume Lample, Myle Ott, Alexis Conneau, Ludovic Denoyer, and Marc{'}Aurelio
  Ranzato. 2018.
\newblock \href {https://doi.org/10.18653/v1/D18-1549} {Phrase-based {\&}
  neural unsupervised machine translation}.
\newblock In \emph{Proceedings of the 2018 Conference on Empirical Methods in
  Natural Language Processing}, pages 5039--5049, Brussels, Belgium.
  Association for Computational Linguistics.

\bibitem[{Ngo et~al.(2020)Ngo, Nguyen, Ha, Dinh, and
  Nguyen}]{ngo-etal-2020-improving}
Thi-Vinh Ngo, Phuong-Thai Nguyen, Thanh-Le Ha, Khac-Quy Dinh, and Le-Minh
  Nguyen. 2020.
\newblock Improving multilingual neural machine translation for low-resource
  languages: {F}rench, {E}nglish - {V}ietnamese.
\newblock In \emph{Proceedings of the 3rd Workshop on Technologies for MT of
  Low Resource Languages}, pages 55--61, Suzhou, China. Association for
  Computational Linguistics.

\bibitem[{Nivre et~al.(2020)Nivre, de~Marneffe, Ginter, Haji{\v{c}}, Manning,
  Pyysalo, Schuster, Tyers, and Zeman}]{universaldeps}
Joakim Nivre, Marie-Catherine de~Marneffe, Filip Ginter, Jan Haji{\v{c}},
  Christopher~D. Manning, Sampo Pyysalo, Sebastian Schuster, Francis Tyers, and
  Daniel Zeman. 2020.
\newblock {U}niversal {D}ependencies v2: An evergrowing multilingual treebank
  collection.
\newblock In \emph{Proceedings of the 12th Language Resources and Evaluation
  Conference}, pages 4034--4043, Marseille, France. European Language Resources
  Association.

\bibitem[{Papineni et~al.(2002)Papineni, Roukos, Ward, and
  Zhu}]{papineni-etal-2002-bleu}
Kishore Papineni, Salim Roukos, Todd Ward, and Wei-Jing Zhu. 2002.
\newblock \href {https://doi.org/10.3115/1073083.1073135} {{B}leu: a method for
  automatic evaluation of machine translation}.
\newblock In \emph{Proceedings of the 40th Annual Meeting of the Association
  for Computational Linguistics}, pages 311--318, Philadelphia, Pennsylvania,
  USA. Association for Computational Linguistics.

\bibitem[{Popovi{\'c}(2015)}]{popovic-2015-chrf}
Maja Popovi{\'c}. 2015.
\newblock \href {https://doi.org/10.18653/v1/W15-3049} {chr{F}: character
  n-gram {F}-score for automatic {MT} evaluation}.
\newblock In \emph{Proceedings of the Tenth Workshop on Statistical Machine
  Translation}, pages 392--395, Lisbon, Portugal. Association for Computational
  Linguistics.

\bibitem[{Popovi{\'{c}} et~al.(2020)Popovi{\'{c}}, Poncelas, Brkic, and
  Way}]{popovic-etal-2020-neural}
Maja Popovi{\'{c}}, Alberto Poncelas, Marija Brkic, and Andy Way. 2020.
\newblock \href {https://www.aclweb.org/anthology/2020.vardial-1.10} {{Neural
  Machine Translation for translating into {C}roatian and {S}erbian}}.
\newblock In \emph{Proceedings of the 7th Workshop on NLP for Similar
  Languages, Varieties and Dialects}, pages 102--113, Barcelona, Spain
  (Online). International Committee on Computational Linguistics (ICCL).

\bibitem[{Post(2018)}]{post-2018-call}
Matt Post. 2018.
\newblock \href {https://doi.org/10.18653/v1/W18-6319} {A call for clarity in
  reporting {BLEU} scores}.
\newblock In \emph{Proceedings of the Third Conference on Machine Translation:
  Research Papers}, pages 186--191, Brussels, Belgium. Association for
  Computational Linguistics.

\bibitem[{Rikters et~al.(2018)Rikters, Pinnis, and
  Kri{\v{s}}lauks}]{rikters-etal-2018-training}
Mat\=iss Rikters, M\=arcis Pinnis, and Rihards Kri{\v{s}}lauks. 2018.
\newblock \href {https://www.aclweb.org/anthology/L18-1595} {{Training and
  Adapting Multilingual {NMT} for Less-resourced and Morphologically Rich
  Languages}}.
\newblock In \emph{Proceedings of the Eleventh International Conference on
  Language Resources and Evaluation ({LREC} 2018)}, Miyazaki, Japan. European
  Language Resources Association (ELRA).

\bibitem[{Sennrich and Haddow(2016)}]{sennrich-haddow-2016-linguistic}
Rico Sennrich and Barry Haddow. 2016.
\newblock \href {https://doi.org/10.18653/v1/W16-2209} {Linguistic input
  features improve neural machine translation}.
\newblock In \emph{Proceedings of the First Conference on Machine Translation:
  Volume 1, Research Papers}, pages 83--91, Berlin, Germany. Association for
  Computational Linguistics.

\bibitem[{Sennrich et~al.(2016)Sennrich, Haddow, and Birch}]{sennrich-bt}
Rico Sennrich, Barry Haddow, and Alexandra Birch. 2016.
\newblock \href {https://doi.org/10.18653/v1/P16-1009} {Improving neural
  machine translation models with monolingual data}.
\newblock In \emph{Proceedings of the 54th Annual Meeting of the Association
  for Computational Linguistics (Volume 1: Long Papers)}, pages 86--96, Berlin,
  Germany. Association for Computational Linguistics.

\bibitem[{Tars and Fishel(2018)}]{tars-fishel-2018}
Sander Tars and M.~Fishel. 2018.
\newblock Multi-domain neural machine translation.
\newblock \emph{ArXiv}, abs/1805.02282.

\bibitem[{Tiedemann(2016)}]{tiedemann-2016-opus}
J{\"o}rg Tiedemann. 2016.
\newblock {OPUS} {--} parallel corpora for everyone.
\newblock In \emph{Proceedings of the 19th Annual Conference of the European
  Association for Machine Translation: Projects/Products}, Riga, Latvia. Baltic
  Journal of Modern Computing.

\bibitem[{Tiedemann(2012)}]{TIEDEMANN12.463}
Jörg Tiedemann. 2012.
\newblock Parallel data, tools and interfaces in opus.
\newblock In \emph{Proceedings of the Eight International Conference on
  Language Resources and Evaluation (LREC'12)}, Istanbul, Turkey. European
  Language Resources Association (ELRA).

\bibitem[{Vaswani et~al.(2017)Vaswani, Shazeer, Parmar, Uszkoreit, Jones,
  Gomez, Kaiser, and Polosukhin}]{transformer-vaswani}
Ashish Vaswani, Noam Shazeer, Niki Parmar, Jakob Uszkoreit, Llion Jones,
  Aidan~N Gomez, \L~ukasz Kaiser, and Illia Polosukhin. 2017.
\newblock Attention is all you need.
\newblock In \emph{Advances in Neural Information Processing Systems},
  volume~30, pages 5998--6008. Curran Associates, Inc.

\bibitem[{Zhang et~al.(2020)Zhang, Williams, Titov, and
  Sennrich}]{zhang-etal-2020-improving}
Biao Zhang, Philip Williams, Ivan Titov, and Rico Sennrich. 2020.
\newblock \href {https://doi.org/10.18653/v1/2020.acl-main.148} {Improving
  massively multilingual neural machine translation and zero-shot translation}.
\newblock In \emph{Proceedings of the 58th Annual Meeting of the Association
  for Computational Linguistics}, pages 1628--1639, Online. Association for
  Computational Linguistics.

\end{thebibliography}

\clearpage

\FloatBarrier
\begin{table*}[h]
{\Large \bf Appendix}
\centering
\begin{adjustbox}{max width=1.0\textwidth}
\begin{tabular}{ l l }
\textbf{\huge A} & \\
\hline
Source & Ja ühe hea külje leiab siidritegija uue nime juures veel. \\
\hline
Baseline & Ja üte hää ütest põlvõst inemiisist lõpõ nime man vahtsõ nime man. \\
ML & Ja üte hää küle om siidritegijä vahtsõ nime man viil. \\
+BT1+FT1(*) & Ja üte hää küle löüd siidritegijä vahtsõ nime man viil. \\
+BT1+BT2(*) & Ja üte hää küle lövvüs siidritegijä vahtsõ nime man viil. \\
\hline
Reference & Ja üte hää küle löüd siidritegijä vahtsõ nime man viil. \\
\hline
\\
\hline
Source & Samas \textbf{tegutsevad kotkad} kultuurmaastikul, mis tähendab, et ka inimesel on tähtis roll selles, et neil hästi läheks. \\
\hline
Baseline & Samal aol om mi kotustõ perändüskultuurmaastikkõ, miä tähendäs, et inemisel tähtsä om tähtsä, et näil olõ-i tähtsä. \\
ML & Saman \textcolor{violet}{omma’ kotka’} kultuurmaastikul, miä tähendäs, et ka inemisel om tähtsä roll tan, et näil häste lääsi. \\
+BT1+FT1(*) & Samal aol \textcolor{violet}{omma kotka} kultuurmaastikul, miä tähendäs, et ka inemisel om tähtsä roll tuun, et näil häste lääsi. \\
+BT1+BT2(*) & Saman toimõndasõq kotkaq kultuurmaastikul, miä tähendäs, et ka inemisel om tähtsä roll tuun, et näil häste lääsiq. \\
\hline
Reference & Saman \textbf{toimõndasõq kotkaq} kultuurmaastikul, miä tähendäs, et ka inemisel om tähtsä roll tuu man, et näil häste lännüq.\\
\hline
\\
\hline
Source & \textbf{Leevakul} elab \textbf{ametlikult pea} 300 inimest. \\
\hline
Baseline & Leevälapjo eläs tähtsä pää inemist. \\
ML & \textcolor{blue}{Leväkul} eläs \textcolor{blue}{virallisesti pää} 300 inemist. \\
+BT1+FT1(*) & Leevakul eläs \textcolor{blue}{virallisesti pää} 300 inemist. \\
+BT1+BT2(*) & Leevakul eläs \textcolor{blue}{virallisesti} pia 300 inemist. \\
\hline
Reference & \textbf{Leevakul} eläs \textbf{kirjo perrä pia} 300 inemist. \\
\hline
\\
\textbf{\huge B} &\\
 \hline
Source & A edesi ei näeq tükk aigo tii pääl üttegi võrokiilset silti.\\
\hline
Baseline & Aga edasi ei näinud tükk aega, tee ühtegi saaklooma ära.\\
ML & Aga edasi ei näe tükk aega tee peal ühtegi võrukeelset ikka. \\
+BT1+FT1 & Aga edasi ei näe tükk aega tee peal ühtegi võrukeelset silti. \\
+BT1\&2+FT1\&2(*) & Aga edasi ei näe tükk aega teel ühtegi võrukeelset silti. \\
\hline
Reference & Aga edasi ei näe tee peal tükk aega ühtegi võrukeelset silti. \\
\hline
 \\
 \hline
 Source & Ütelt puult tulõ hoita vannu mõtsu, et nä saanu ummi \textbf{pessi} kohegi ehitä. \\
 \hline
 Baseline & Ühis poolt tuleb hoida vanade metsade, et nad saanud märkimisväärset kuhugi ehitanud. \\
 ML & Ühel pool tuleb hoida vana metsa, et nad saaksid oma \textcolor{violet}{pesu} ehitada. \\
+BT1+FT1 & Ühel pool tuleb hoida vana metsa, et nad saaksid oma \textcolor{violet}{pessi} kuhugi ehitada. \\
+BT1\&2+FT1\&2(*) & ühelt poolt tuleb hoida vanu metsasid, et nad saaksid oma \textcolor{violet}{pessi} kuhugi ehitada. \\
 \hline
 Reference & Ühelt poolt tuleb hoida vanu metsi, et nad saaks oma \textbf{pesasid} kuhugi ehitada. \\
 \hline
\\
\hline
 Source & Nii om võimalus telefon võita ka Uma Lehe \textbf{teljäl}.\\
\hline
Baseline & Nii on võimalus telefongu ka Uma Pidoga mitmeti seotud. \\
ML & Nii on võimalus telefon võita ka Uma Lehe \textcolor{blue}{telgis}. \\
+BT1+FT1 & Nii on võimalus telefon võita ka Uma Lehe \textcolor{blue}{telgil}. \\
+BT1\&2+FT1\&2(*) & Nii on võimalus telefon võita ka Uma Lehe \textcolor{blue}{telgil} . \\
\hline
Reference & Nii on võimalus telefon võita ka Uma Lehe \textbf{tellijal}. \\
\hline
\\
\textbf{\huge C} & \\
\hline
Source & Uutta nimeä \textbf{keksiessä} oli tärkeää, että olisi selkeä yhteys paikalliseen yhteisöön ja että nimi auttaisi kertomaan yrityksen tarinan. \\
\hline
Baseline & M uhccin muitalin lei dehálaš, ahte livččii čielga oktavuohta báikkálaš servodahkii ja ahte namma veahkehivččii muitalit lihkastagaide. \\
ML & Ođđa namma lei dehálaš, ahte livččii čielga oktavuohta báikkálaš servošii ja ahte namma veahkehivččii muitalit fitnodaga máidnasiid. \\
+BT1 & Ođđa nama \textcolor{blue}{huksemis} lei dehálaš, ahte livččii čielga oktavuohta báikkálaš servodahkii ja ahte namma veahkehivččii muitalit fitnodaga máidnasa. \\
\hline
Reference & Ođđa nama \textbf{hutkkadettiin} lea dehálaš, ahte livčče čielga oktavuohta báikkálaš servvodahkii ja ahte namma veahkehivčče muitalit fitnodaga muitalusa. \\
\hline
\\
\hline
Source & \textbf{Kansa} kokoontuu entiseen \textbf{koulutaloon}, jossa on myös kirjasto. \\
\hline
Baseline & Riikkabeaivevahku čoahkkana ságadoalliriikkas, mas leat maid girjerájus.\\
ML & \textcolor{violet}{Álbmoga} čoahkkana ovddeš \textcolor{violet}{skuvllas}, mas lea maid girjerádju. \\
+BT1 & Álbmot čoahkkana ovddeš \textcolor{violet}{skuvladássái}, mas lea maid girjerádju. \\
\hline
Reference & \textbf{Álbmot} čoahkkana boares \textbf{skuvlavistái}, mas lea maiddái girjerádju. \\
\hline
\\
\hline
Source & Sosiaalisessa mediassa \textbf{pitivät} ihmiset eniten tehtävästä “Puhu tai postaa \textbf{yksi vitsi} tai tarina \textbf{võron kielellä}”. \\
\hline
Baseline & Sosiála medias atne olbmot eanemus barggus “ Ominayak oktavuođaváldimiid dehe máidnumaõjjstõõllâmǩe'rjj lea oaivvilduvvon. \\
ML & Sosiála medias \textcolor{blue}{doalai} olbmuid eanemus bargguin ”Puhu dahje poasta \textcolor{blue}{okta nja} dahje máidnasa \textcolor{violet}{gillii}. \\
+BT1 & Sosiála medias \textcolor{blue}{dolle} olbmot eanemusat bargguin “Puhu dahje poasta \textcolor{blue}{okta njaš} dahje máidnasa \textcolor{violet}{vuonagillii}”. \\
\hline
Reference & Sosiála medias \textbf{liikojedje} olbmot maiddái bargobihtás “Muital dahje postte \textbf{ovtta cukcasa} dahje máidnasa \textbf{võro gillii}.” \\
\hline
\end{tabular}
\end{adjustbox}
\caption{Translation examples. A: Estonian-Võro, B: Võro-Estonian, C: Finnish-North Saami. \textcolor{blue}{blue} - incorrect word, \textcolor{violet}{violet} - incorrect form/case/tense or partially incorrect.}
\label{table:5}
\end{table*}
\FloatBarrier

\begin{table*}[t]
\centering
\begin{adjustbox}{max width=1.0\textwidth}
\begin{tabular}{ l|c|c|c|c|c|c|c|c|c|c|c } 
 Model & et-fi & fi-et & et-vro & vro-et & fi-sme & sme-fi & fi-sma & sma-fi & sme-sma & sma-sme & $CHRF_{low}$  \\
 \hline\hline
 Baselines & \textbf{0.602} & 0.573 & 0.353 & 0.390 & 0.577 & 0.577 & 0.282 & 0.274 & 0.330 & 0.301 & 0.385\\
 \hline\hline
 Multilingual (ML) & 0.595 & 0.578 & 0.510 & 0.540 & 0.631 & 0.650 & 0.376 & 0.348 & 0.546 & 0.525 & 0.516\\
 \hline\hline
 + BT1 & 0.600 & 0.583 & 0.531 & 0.551 & 0.639 & 0.659 & 0.408 & 0.348 & 0.557 & 0.531 & 0.528\\ 
 + BT1(*) & 0.592 & 0.575 & 0.526 &	0.556 & 0.639 & 0.659 & 0.420 & 0.349 & 0.566 & 0.532 & 0.531\\
 + BT1 + FT1 & 0.595 & 0.584 & 0.526 & 0.558 & 0.634 & 0.654 & 0.369 & 0.353 & 0.544 & 0.533 & 0.525\\
 + BT1 + FT1(*) & 0.596	& 0.575 & 0.537	& 0.558 & 0.636 & 0.656 & 0.392 & 0.349 & 0.551 & 0.531 & 0.525\\
 \hline\hline
 + BT2 & 0.598 & \textbf{0.585} & 0.535 & 0.560 & \textbf{0.640} & 0.663 & 0.418 & 0.358 & 0.563 & 0.537 & 0.534\\
 + BT1 + BT2(*) & 0.595 & 0.583 & \textbf{0.539} & \textbf{0.565} & 0.636 & 0.663 & \textbf{0.436} & \textbf{0.364} & \textbf{0.569} & \textbf{0.539} & \textbf{0.539}\\
 + BT1 + BT2(**) & 0.594 & 0.579 & 0.530 & 0.563 & \textbf{0.640} & \textbf{0.665} & 0.423 & 0.354 & 0.567 & 0.536 & 0.535\\
 + BT1\&2 + FT1\&2(*) & 0.592 &	0.578 &	0.530 & \textbf{0.565} & 0.634 & 0.662 & 0.399 & 0.350 & 0.564 & \textbf{0.539} & 0.530\\
 \hline\hline
 BT1 & 0.526 & 0.515 & 0.480 & 0.523 & 0.582 & 0.607 & 0.371 & 0.296 & 0.524 & 0.488 & 0.484\\
 BT1(*) & 0.349 & 0.356 & 0.455 & 0.473 & 0.459 & 0.443 & 0.348 & 0.253 & 0.488 & 0.402 & 0.415\\
\end{tabular}
\end{adjustbox}
\caption{CHRF scores. (*) - trained without pre-trained weights, (**) - trained on \textit{+ BT1(*)} weights. \textit{BT} - back-translation data set, \textit{FT} - forward-translation data set, $CHRF_{low}$ - average CHRF score on low-resource language pairs (excluding ET-FI and FI-ET), \textbf{bold} - best CHRF score for a language pair.}
\label{table:9}
\end{table*}

\end{document}